\pdfoutput=1
\def\cvprPaperID{4521}
\def\confName{CVPR}
\def\confYear{2022}

\def\paperTitle{Simple but Effective: CLIP Embeddings for Embodied AI}

\def\authorBlock{
    Apoorv Khandelwal\thanks{Equal contribution} \qquad
    Luca Weihs\footnotemark[1] \qquad
    Roozbeh Mottaghi \qquad
    Aniruddha Kembhavi \\
    Allen Institute for AI \\
    {\tt\small \{apoorvk, lucaw, roozbehm, anik\}@allenai.org}
}

\newif\ifreview 
\newif\ifarxiv \newcommand{\arxiv}{\arxivtrue}
\newif\ifcamera 
\newif\ifrebuttal 

\arxiv %

\documentclass[10pt,twocolumn,letterpaper]{article}
\ifreview \usepackage[review]{cvpr} \fi
\ifarxiv \usepackage[pagenumbers]{cvpr} \fi
\ifrebuttal \usepackage[rebuttal]{cvpr} \fi
\ifcamera \usepackage{cvpr} \fi

\usepackage{graphicx}
\usepackage{amsmath}
\usepackage{amssymb}
\usepackage{booktabs}

\usepackage{times}
\usepackage{epsfig}
\usepackage[table,xcdraw]{xcolor}
\usepackage{caption}
\usepackage{float}
\usepackage{placeins}
\usepackage{booktabs}
\usepackage{color, colortbl}
\usepackage{stfloats}
\usepackage{enumitem}
\usepackage{tabularx}
\usepackage{multirow}
\usepackage{xspace}
\usepackage{url}
\usepackage{subcaption}
\usepackage{xcolor}
\usepackage[hang,flushmargin]{footmisc}

\ifcamera \usepackage[accsupp]{axessibility} \fi

\newcommand{\modelclip}{EmbCLIP\xspace}
\definecolor{ColorTableUs}{HTML}{FFF2CC}

\newcommand{\thor}{\textsc{AI2-THOR}\xspace}
\newcommand{\pnav}{\textsc{PointNav}\xspace}
\newcommand{\onav}{\textsc{ObjectNav}\xspace}
\newcommand{\roomr}{\textsc{RoomR}\xspace}

\definecolor{clp}{rgb}{0.92,0.49,0.19}
\definecolor{img}{rgb}{0.26,0.44,0.76}

\newcommand{\parbf}[1]{{\noindent \textbf{#1.}}}

\newcommand{\supp}{supplemental material\xspace}
\ifarxiv \renewcommand{\supp}{appendix\xspace} \fi

\usepackage{xr-hyper}

\makeatletter
\newcommand*{\addFileDependency}[1]{
  \typeout{(#1)}
  \@addtofilelist{#1}
  \IfFileExists{#1}{}{\typeout{No file #1.}}
}

\makeatother

\usepackage[pagebackref,breaklinks,colorlinks]{hyperref}
\usepackage[capitalize]{cleveref}
\crefname{section}{Sec.}{Secs.}
\crefname{table}{Table}{Tables}
\crefname{figure}{Fig.}{Figs.}

\begin{document}
\title{\paperTitle}
\author{\authorBlock}
\maketitle
\begin{abstract}
Contrastive language image pretraining (CLIP) encoders have been shown to be beneficial for a range of visual tasks from classification and detection to captioning and image manipulation. We investigate the effectiveness of CLIP visual backbones for Embodied AI tasks. We build incredibly simple baselines, named \modelclip, with no task specific architectures, inductive biases (such as the use of semantic maps), auxiliary tasks during training, or depth maps---yet we find that our improved baselines perform very well across a range of tasks and simulators. \modelclip tops the RoboTHOR ObjectNav leaderboard by a huge margin of 20 pts (Success Rate). It tops the iTHOR 1-Phase Rearrangement leaderboard, beating the next best submission, which employs Active Neural Mapping, and more than doubling the \% Fixed Strict metric (0.08 to 0.17). It also beats the winners of the 2021 Habitat ObjectNav Challenge, which employ auxiliary tasks, depth maps, and human demonstrations, and those of the 2019 Habitat PointNav Challenge. We evaluate the ability of CLIP's visual representations at capturing semantic information about input observations---primitives that are useful for navigation-heavy embodied tasks---and find that CLIP's representations encode these primitives more effectively than ImageNet-pretrained backbones. Finally, we extend one of our baselines, producing an agent capable of zero-shot object navigation that can navigate to objects that were not used as targets during training. Our code and models are available at \url{https://github.com/allenai/embodied-clip}.
\end{abstract}

\section{Introduction}

\begin{figure}[tp]
    \centering
    \includegraphics[width=\linewidth]{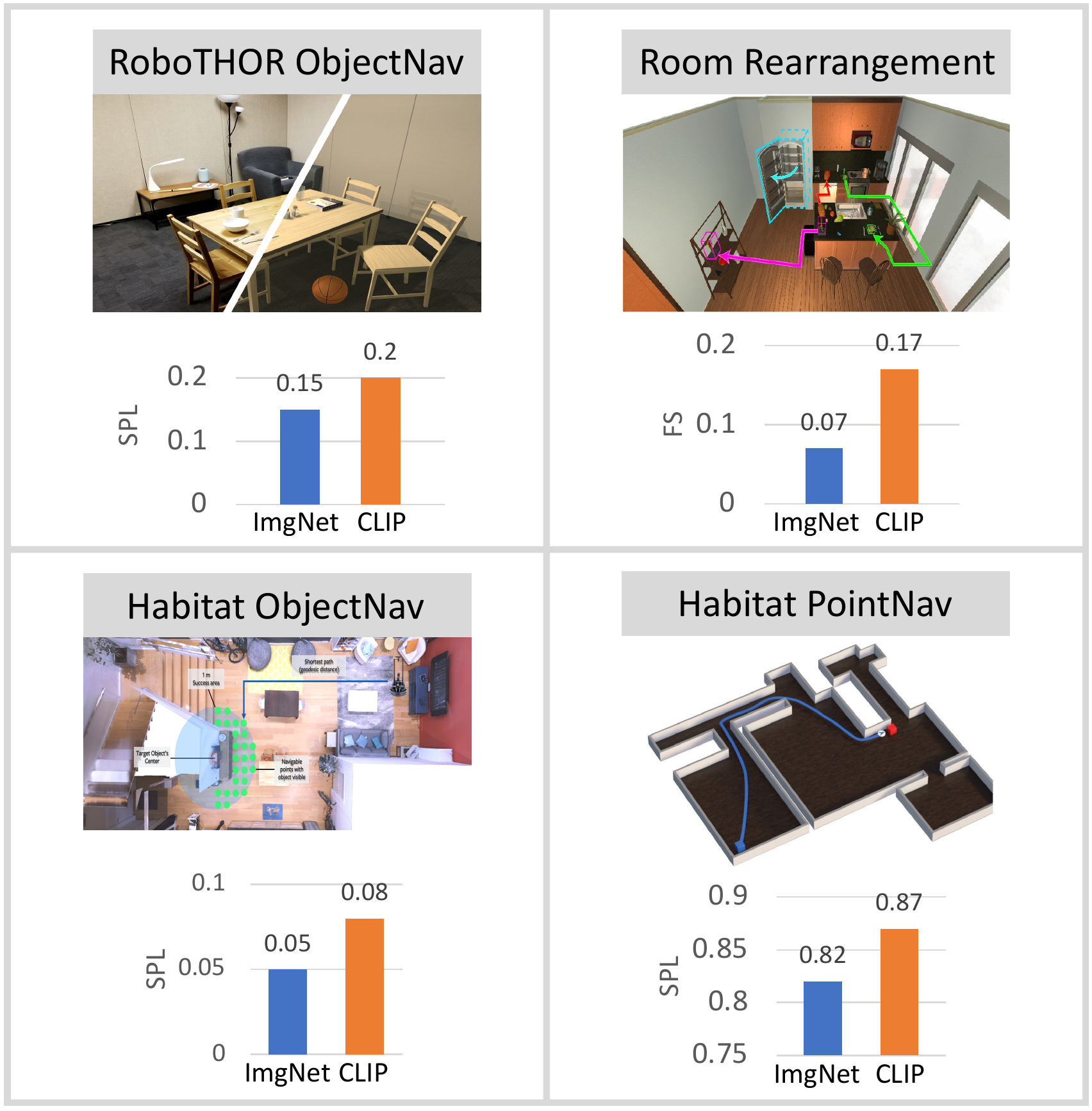}
    \caption{We show powerful visual encoders are important for Embodied AI tasks. We consider four navigation-heavy tasks and show {\textcolor{clp}{CLIP}}-based encoders provide massive gains over ResNet architectures trained on {\textcolor{img}{ImageNet}}. More interestingly, using only RGB images as input, they outperform approaches employing depth images, maps, and more sophisticated architectures.}

    \label{fig:teaser}
\end{figure}

\label{sec:intro}

The CLIP family of neural networks have produced very impressive results at a series of visual recognition tasks, including an astonishing zero-shot performance on ImageNet that matches the accuracy of a fully supervised ResNet-50 model~\cite{clip}. Unsurprisingly, visual representations provided by CLIP have now also been shown to provide improvements across other computer vision tasks such as open vocabulary object detection~\cite{gu2021openvocabulary}, image captioning, visual question answering, and visual entailment~\cite{Shen2021HowMC}. In this work, we investigate the effectiveness of CLIP's visual representations at tasks in the domain of Embodied AI.

Embodied AI tasks involve agents that learn to navigate and interact with their environments. Shen \emph{et al.}~\cite{Shen2021HowMC} demonstrated that CLIP features can provide gains in instruction following along a navigation graph: a task where an agent traverses the graph using linguistic instructions such as \emph{Walk past the piano}. Building upon this outcome, we provide a thorough investigation of CLIP-powered Embodied AI models at tasks that require agents to use low level instructions (such as \texttt{Move Ahead}, \texttt{Turn}, \texttt{Look Down}, and \texttt{Pick Up}) to take steps in a scene and interact with objects. Such tasks, including Object Goal Navigation~\cite{batra2020objectnav} and Room Rearrangement~\cite{Batra2020RearrangementAC}, are inherently navigation-heavy. As a result, visual representations of observations in such tasks must not just be effective for recognizing categories, but also at encoding primitives such as walkable surfaces, free space, surface normals, and geometric structures.

We build a series of simple baselines using the CLIP ResNet-50 visual encoder. Our embodied agents input CLIP representations of observations that are frozen and not fine-tuned for the task at hand. These representations are combined with the goal specification and then passed into a recurrent neural network (RNN) unit to provide memory. A linear layer transforms the hidden activations of the RNN to a distribution over the agent's actions. These simple baselines have very few task specific design choices, do not use depth maps (which are commonly regarded as critical for good results), use no spatial or semantic maps, and employ no auxiliary tasks during training. And yet, without these design choices that have been empirically proven to be effective, our simple CLIP-based baselines are very effective---vaulting to the top of two leaderboards for the tasks of Object Goal Navigation and Room Rearrangement in the \thor environment, and outperforming the Habitat simulator challenge winners for Object Goal Navigation in 2021 and Point Goal Navigation in 2019.

These results are surprising and suggest that CLIP's visual representations are powerful and encode visual primitives that are useful to downstream Embodied AI tasks. Indeed, we find that CLIP's representations outperform those from ImageNet pretraining by using a linear probe at four primitive tasks: object presence, object localization, reachability and free space estimation. In particular, CLIP's representations provide a +3 point absolute improvement on three of our probing studies and +6.5 point (+16 \% relative) at the object localization probe.

We analyze 4 visual encoders (2 ResNet models pretrained on ImageNet and 2 ResNet models pretrained with CLIP) and use them to train 4 Object Goal Navigation agents. We then study the correlation between ImageNet Top-1 Accuracy and Success Rate for Object Goal Navigation. We find that ImageNet accuracy alone is not a good indicator of an encoder's suitability for Embodied AI tasks, and that it may be more useful to probe its representations for semantic and geometric information. However, our results do indicate that Embodied AI agents can continue to benefit from better visual encodings.

Finally, we use CLIP's visual and textual encoders in a simple architecture with few learnable parameters to train an agent for zero-shot Object Goal Navigation (\ie navigating to objects that are not targets at training time). Our results are promising: the agent achieves roughly half the Success Rate for unseen objects compared to that for seen objects. They suggest that using visual representations that are strongly grounded in language can help build Embodied AI models that can generalize not only to new scenes, but also to new objects.

\section{Related Work}
\label{sec:related}

\parbf{Embodied AI}
There are now many simulators and tasks in Embodied AI. Simulators are photo-realistic, simulate physics, and enable interaction between agents and their environments~\cite{Kolve2017AI2THORAn, Deitke2020RoboTHORAn, Ehsani2021ManipulaTHORA, habitat19iccv, szot2021habitat, xiazamirhe2018gibsonenv, Shen2020iGibsonAS, Li2021iGibson2O}. Large scene datasets accompany these simulators~\cite{Matterport3D, xiazamirhe2018gibsonenv, Ramakrishnan2021HabitatMatterport3D}. Many creative tasks~\cite{Kolve2017AI2THORAn, Deitke2020RoboTHORAn, Ehsani2021ManipulaTHORA, batra2020objectnav, habitat2020sim2real, ALFRED20, RoomR, Batra2020RearrangementAC, Jain2019TwoBP} have been proposed, such as navigation, rearrangement, and furniture moving, and require complex combinations of skills, including perception, visual reasoning, communication, coordination, and action selection. While fast progress has been made on the task of Point Goal Navigation~\cite{Wijmans2020DDPPOLN} in the presence of ideal sensors, most other tasks remain challenging. 

\parbf{Visual Encoders in Embodied AI}
Due to inefficiency of larger models, most prior work uses lightweight CNN or ResNet models as visual encoders in Embodied AI tasks. This is particularly the case when sample or compute efficiency is of concern. Extensive experimentation has been done to compare the relative advantages of simple CNNs against ResNet-18 for Point Goal Navigation in Habitat~\cite{habitat19iccv} in this low sample/compute regime~\cite{WijmansEtAl2020HowToTrainPointGoal}. Although experimentation with larger ResNet visual encoders for Embodied AI has provided some evidence that such models can marginally improve downstream performance, there is also interesting evidence showing that using ImageNet-pretrained weights can harm downstream transfer to new tasks~\cite{Wijmans2020DDPPOLN} (\eg when using transfer learning to perform exploration with a model trained for Point Goal Navigation).

\parbf{CLIP and CLIP based models}
CLIP's visual representations have proven very effective at zero-shot image recognition (compared to ImageNet-pretrained ResNet representations, when a fully supervised linear classification layer is trained upon them) on benchmarks such as ImageNet, CIFAR100, Kinetics700 and UCF 101~\cite{clip}. These representations are also effective at vision-language tasks such as VQA and Visual Entailment~\cite{Shen2021HowMC}. CLIP's open vocabulary image classification ability has also been distilled into an open vocabulary object detection~\cite{gu2021openvocabulary}, proving to be very effective. CLIP has also proven to be effective at video tasks particularly text based video retrieval~\cite{Luo2021CLIP4ClipAE,PortilloQuintero2021ASF,Fang2021CLIP2VideoMV}. Finally, CLIP has been used for pixel generative tasks, including a text-based interface for StyleGAN image manipulation~\cite{Patashnik2021StyleCLIPTM} and synthesizing novel drawings based on natural language input~\cite{Frans2021CLIPDrawET}. In the Embodied AI domain,  CLIP has been employed with a Transporter~\cite{Zeng2020TransporterNR} network to solve language-specified tabletop tasks such as packing objects and folding cloth~\cite{shridhar2021cliport}. CLIP has also been shown to help in instruction following~\cite{Shen2021HowMC} which requires the agent to move along the edges of a navigation graph. This task requires models to identify objects and locations in the scene via their names and descriptions. Buoyed by this success, we investigate the use of CLIP towards navigation-heavy tasks that require the agent to use atomic actions (such as \texttt{Move Forward}, \texttt{Turn Left}, \texttt{Pick Up}, \texttt{Place}, etc.) to navigate within a scene. To do so successfully, visual encoders must not only identify objects, but also capture an understanding of depth, free space, surface normals, etc. Our experiments show that CLIP is remarkably effective at such tasks.

\begin{figure}[tp]
    \centering
    \includegraphics[width=\linewidth]{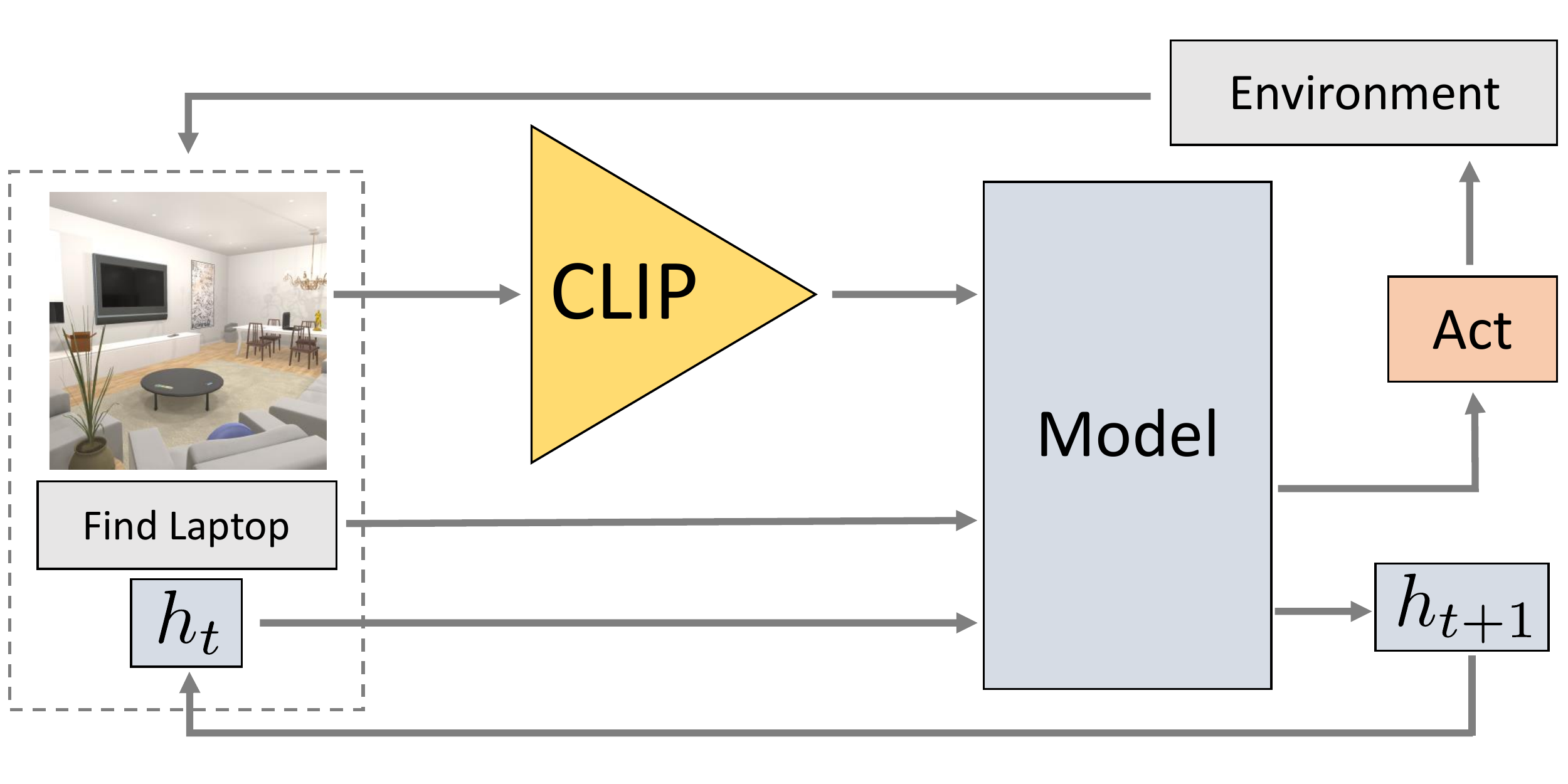}
    \caption{\textbf{Model overview.} All of our models use the same high-level architecture illustrated here, where a model (employing an RNN) receives features from the visual encoder, task definition, and previous hidden units as input, and outputs an action.}

    \vspace{-0.05in}
    \label{fig:model}
\end{figure}

\section{Using CLIP in Embodied AI}
\label{sec:clip}

CLIP~\cite{clip} is a recently released family of image and text encoders that are pretrained to contrast between corresponding and non-corresponding image--caption pairs. Although seemingly simple, this pretraining task, when employed at the scale of 400M image--caption pairs, leads to very powerful visual encoders. CLIP's visual backbones perform very well at a suite of computer vision datasets in a zero-shot setting, including matching the accuracy of the original fully supervised ResNet-50 model on the ImageNet benchmark.

CLIP's visual representations have proven to be effective for several other vision and vision--language tasks (see~\cref{sec:related}). We investigate their effectiveness at four navigation-heavy Embodied AI tasks: Object Goal Navigation (\onav) in RoboTHOR and Habitat, Point Goal Navigation (\pnav) in Habitat, and Room Rearrangement (\roomr) in iTHOR. For each of these benchmarks, we use a baseline implementation provided by its authors and substitute its visual encoder (often a shallow CNN, such as ResNet-18) with CLIP ResNet-50.

The resulting baselines have very similar core architectures. \cref{fig:model} shows the schematic of a general baseline model with a CLIP visual encoder. They primarily differ in their usage of the goal description (object name in \onav, coordinates in \pnav, etc.) and processing of the visual features. Below, we describe this core baseline architecture in the context of \onav in RoboTHOR. Please see the \supp for details on other tasks.

Specifically, the model for \onav in RoboTHOR receives a $3{\times}224{\times}224$ RGB image $i$ and integer $g\in\{0,...,11\}$ indicating the goal object category as input. The RGB image is encoded into a $2048{\times}7{\times}7$ tensor $I$ by a CLIP ResNet-50 model whose weights are frozen and final attention pooling and classification layers have been removed. $g$ is used to index a (trainable) embedding matrix and form a 32-dim goal embedding $G$. A two layer CNN then compresses $I$ to form a tensor $I'$ of shape $32{\times}7{\times}7$. The vector $G$ is tiled to a shape of $32{\times}7{\times}7$ and concatenated with $I'$. This $64{\times}7{\times}7$ tensor is passed through another two-layer CNN (resulting in a $32{\times}7{\times}7$ shape) and flattened to form $V$: a 1568-dim goal-conditioned visual embedding. $V$ is passed into a 1-layer GRU with 512 hidden units, along with any prior hidden state. Then, one linear layer maps the GRU output to a 6-dimensional vector of logits and another linear layer maps it to a scalar, respectively forming the actor (\ie policy) and critic (\ie value) heads commonly used in reinforcement learning.

In this work, we only consider CLIP ResNet-50, but our baselines can be trivially extended for other CLIP variants.

\section{When is CLIP effective?}
\label{sec:when}

We build CLIP-based baselines (using CLIP ResNet-50) across multiple navigation-heavy tasks and across simulators. We compare these models to each task's public leaderboards and, in all cases, find that our models are competitive with SOTA methods---even surpassing the previous best entries on RoboTHOR \onav and iTHOR \roomr by large margins. We present these leaderboards\footnote{\footnotesize As of 11/16/2021} in~\cref{tab:robothor_objectnav,tab:ithor_rearrangement,tab:habitat_objectnav,tab:habitat_pointnav} and have also submitted our model to these online leaderboards (besides the 2019 Habitat \pnav leaderboard, which is now closed).

This is a particularly surprising result, since other methods required significant additional research and engineering efforts (\eg task specific designs or massively distributed training for long durations) compared to our relatively simple modification. Importantly, \textbf{our CLIP baselines only use RGB}, whereas most models on leaderboards also use perfect depth information.

In each of the tasks, we also train a baseline agent with an ImageNet-pretrained ResNet-50. This ensures a fair comparison to the CLIP-based baseline w.r.t. parameter counts. We evaluate the checkpoint for each agent with the best validation set performance. In all four experiments, we find that pretraining with CLIP provides significant gains in success metrics compared to on ImageNet.

 Hereafter, we refer to these baselines (with frozen ResNet-50 visual encoders and RGB as the only visual input) as our ``CLIP agent'' and ``ImageNet agent'' (by their method of pretraining).

\subsection{Object Goal Navigation in RoboTHOR}
\label{ssec:robothor_objectnav}

\begin{table}[t]
\centering
\resizebox{\columnwidth}{!}{%

\begin{tabular}{lcccc}
\toprule
Models & SPL & SR & SPL Prox & SR Prox \\
\midrule
\rowcolor{ColorTableUs}
ResNet-50 (CLIP) & \textbf{0.20} & \textbf{0.47} & \textbf{0.20} & \textbf{0.48} \\
ResNet-50 (ImageNet) & 0.15 & 0.34 & 0.15 & 0.35 \\
\midrule \midrule
\rowcolor{ColorTableUs}
(1) \modelclip (Ours) & \textbf{0.20} & \textbf{0.47} & \textbf{0.20} & \textbf{0.48} \\
(2) Action Boost & 0.12 & 0.28 & 0.12 & 0.30 \\
(3) RGB+D ResNet18 & 0.11 & 0.26 & 0.12 & 0.28 \\
\multicolumn{5}{c}{$\cdots$} \\
\bottomrule
\end{tabular}

}
\caption{
\textbf{RoboTHOR \onav Challenge (2021).} Here, we show test success metrics: Success weighted by Path Length (SPL) and Success Rate (SR). We compare our baseline agents (above) and against leaderboard entries (below).
}
\vspace{-0.05in}
\label{tab:robothor_objectnav}
\end{table}

\parbf{Task}
\onav requires an agent to navigate through its environment and find an object of a given category. An instance of the \onav task has been developed for the simulated RoboTHOR environment~\cite{Deitke2020RoboTHORAn}. In RoboTHOR \onav, a robotic agent\footnote{\footnotesize Based on the LoCoBot robot (\href{http://www.locobot.org/}{\texttt{www.locobot.org}})} is placed into a near-photorealistic home environment at a random location. It is then given one of twelve goal object categories (\eg apple) and must navigate to this object using \texttt{MoveAhead}, \texttt{RotateRight}, \texttt{RotateLeft}, \texttt{LookUp}, and \texttt{LookDown} actions. The agent is successful if it takes a special \texttt{Done} action and there is an instance of the goal object category that is both visible to the agent and within 1m of the agent.

\parbf{Prior work}
There have been two challenges for RoboTHOR \onav (in 2020 and 2021), as well as several prior works that have used RoboTHOR \onav as a testbed for reinforcement and imitation learning methodologies~\cite{JainEtAl2021GridToPix,WeihsEtAl2021Advisor} and for studying model robustness~\cite{PrithvijitEtAl2021RobustNav}. We present results on the 2021 RoboTHOR \onav leaderboard\footnote{\footnotesize\href{https:leaderboard.allenai.org/robothor_objectnav}{\texttt{leaderboard.allenai.org/robothor\_objectnav}}}. The best performing models prior to our submission are ``Action Boost''~\cite{robothoractionboost} which obtains a Test Set SPL of 0.12 and 	
``RGB+D ResNet18'' which obtains a Test Set SPL of 0.11 with code available via AllenAct~\cite{AllenAct}.

\parbf{Performance}
We modify the ``RGB+D ResNet18'' baseline model for our experiment (as described in~\cref{sec:clip}) and train our ImageNet and CLIP agents for 200M environment steps using DD-PPO. In~\cref{tab:robothor_objectnav}, we display the performance of our models. Our CLIP agent dramatically outperforms both our ImageNet agent (by 1.3--1.4x in SPL and Success Rate) and the ``Action Boost'' model (by 1.7x in SPL and Success Rate).

\subsection{Room Rearrangement in iTHOR}
\label{ssec:ithor_rearrangement}

\begin{table}[t]
\centering

\begin{tabular}{lcccc}
\toprule
Model & FS & SR & E & M \\
\midrule
\rowcolor{ColorTableUs}
ResNet-50 (CLIP) & \textbf{0.17} & \textbf{0.08} & \textbf{0.89} & 0.88 \\
ResNet-50 (ImageNet) & 0.07 & 0.03 & 1.06 & 1.05 \\
\midrule \midrule
\rowcolor{ColorTableUs}
(1) \modelclip (Ours) & \textbf{0.17} & \textbf{0.08} & \textbf{0.89} & 0.88 \\
(2) RN18 + ANM IL~\cite{RoomR} & 0.09 & 0.03 & 1.04 & 1.05 \\
(3) RN18 + IL~\cite{RoomR} & 0.06 & 0.03 & 1.09 & 1.11 \\
\multicolumn{5}{c}{$\cdots$} \\
\bottomrule
\end{tabular}

\caption{
\textbf{iTHOR 1-Phase Rearrangement Challenge (2021).} We report test set metrics for both our baselines (above) and leaderboard entries (below).
}
\vspace{-0.05in}
\label{tab:ithor_rearrangement}
\end{table}

\parbf{Task}
A consortium of researchers in Embodied AI have recently proposed for Room Rearrangement to be the next frontier challenge~\cite{Batra2020RearrangementAC} for existing models and methods. An instance of the \roomr task has been developed for the simulated iTHOR environment~\cite{RoomR}. This proposed task has 1-phase and 2-phase variants, and we focus solely on the 1-phase setting in this work. In this 1-phase task, the agent is placed into a room and is given two images at every step. One image depicts the room ``as it is'' from the agent's egocentric viewpoint. The other image (from the same perspective) shows the room ``as it should be''---namely, with several objects in different locations or in different states (\eg a laptop may be on a table rather than on a sofa, or a cabinet may be open rather than closed). Using a discrete set of standard navigational actions (\eg \texttt{MoveAhead}, \texttt{RotateRight}, \etc) and higher-level semantic actions (\eg \texttt{PickUpX} for an object category \texttt{X}), the agent must rearrange the objects in the environment from their current states to their ``as it should be'' states.

\parbf{Prior work}
We report results for the 2021 iTHOR Rearrangement Challenge\footnote{\href{https://leaderboard.allenai.org/ithor_rearrangement_1phase}{\texttt{leaderboard.allenai.org/ithor\_rearrangement\_1phase}}}.
The best performing model prior to our submission is the ``RN18+ANM IL'' model, which used a semantic mapping module based on the Active Neural SLAM (ANM)~\cite{chaplot2020learning} architecture (proposed by Chaplot \etal) and was trained via imitation learning. The second best performing method (proposed by the iTHOR Rearrangement authors), which we adopt, uses an ImageNet-pretrained ResNet-18 visual encoder with frozen weights, an attention module to compare image pairs, and a 1-layer GRU to integrate observations over time.

\parbf{Performance}
Rather than relying on complex inductive biases in the form of semantic maps, we base our architecture on the second best performing method (``RN18 + IL'' in~\cref{tab:ithor_rearrangement}). We train the ImageNet and CLIP agents in our experiment using imitation learning for 70 million steps. The performances of our models and the previous two best baselines are shown in~\cref{tab:ithor_rearrangement}. The results are dramatic: our CLIP agent achieves a 1.94x improvement over the ANM model (from 9\% to 17\%) in the \texttt{FixedStrict} (FS) metric used to rank \roomr models. On the other hand, our ImageNet agent (7\% FS) with a ResNet-50 encoder performs (1) only slightly better than the ``RN18 + IL'' baseline (6\% FS), which it was based on, and (2) worse than the ``RN18 + ANM IL'' model (9\% FS). This demonstrates that only making the visual encoder deeper is relatively ineffective. Beating the ANM-based model is somewhat remarkable, demonstrating that using CLIP representations offers a far greater improvement than even the inductive bias of a map.

\subsection{Object Goal Navigation in Habitat}
\label{ssec:objectnav_habitat}

\begin{table}[t]
\centering
\resizebox{\columnwidth}{!}{%

\begin{tabular}{lcccc}
\toprule
Models & SPL & SR & SoftSPL & Goal Dist \\
\midrule
\rowcolor{ColorTableUs}
ResNet-50 (CLIP) & \textbf{0.08} & \textbf{0.18} & \textbf{0.20} & \textbf{7.92} \\
ResNet-50 (ImageNet) & 0.05 & 0.13 & 0.17 & 8.69 \\
\midrule \midrule
(1) yuumi\_the\_magic\_cat~\cite{luo2022stubborn} & \textbf{0.10} & 0.22 & 0.18 & 9.17 \\
(2) TreasureHunt~\cite{Maksymets_2021_ICCV} & 0.09 & 0.21 & 0.17 & 9.20 \\
(3) Habitat on Web (IL-HD)~\cite{habitatweb} & 0.08 & \textbf{0.24} & 0.16 & \textbf{7.88} \\
\rowcolor{ColorTableUs}
(4) \modelclip (Ours) & 0.08 & 0.18 & \textbf{0.20} & 7.92 \\
(-) Habitat on Web\textsuperscript{2021}~\cite{habitatweb} & 0.07 & 0.21 & 0.15 & 8.26 \\
(5) Red Rabbit\textsuperscript{2021}~\cite{Ye_2021_ICCV} & 0.06 & 0.24 & 0.12 & 9.15 \\
\multicolumn{1}{c}{$\cdots$} & \multicolumn{4}{c}{} \\
(9) DD-PPO & 0.00 & 0.00 & 0.01 & 10.326 \\
\bottomrule
\end{tabular}

}
\caption{
\textbf{Habitat \onav Challenge (2021).} We report Test-Standard split metrics for both our baselines (above) and leaderboard entries (below). Entries with the superscript 2021 indicate challenge winners.
}
\vspace{-0.05in}
\label{tab:habitat_objectnav}
\end{table}

\parbf{Task}
\onav in Habitat is defined similarly to the task in RoboTHOR (\cref{ssec:robothor_objectnav}), including having the same action space and robotic agent design. However, Habitat uses 21 objects and does not require the agent to be looking at a target object for the episode to be successful. For the \onav challenge, Habitat uses scenes from the MatterPort3D~\cite{Matterport3D} dataset of real-world indoor spaces.

\parbf{Prior work}
\onav has been a part of the 2020 and 2021 Habitat challenges. However, in the 2021 leaderboard\footnote{\footnotesize\href{https://eval.ai/web/challenges/challenge-page/802/leaderboard/2195}{\texttt{eval.ai/web/challenges/challenge-page\\/802/leaderboard/2195}}}, the baseline (an RGB-D agent with a ResNet-50 visual encoder trained from scratch using DD-PPO) scores 0.00 SPL on the test-standard set---this same method was shown to ``solve'' \pnav in Habitat~\cite{Wijmans2020DDPPOLN} (and we explore this setting further in~\cref{ssec:habitat_pointnav}). As even the current leader (after over two years of entries) only achieves 0.10 SPL, we can see that this task remains quite challenging. As the baseline is mapless and has task-agnostic components, competitive leaderboard entries include numerous modifications to increase performance (\eg augmenting scenes with 3D object scans~\cite{Maksymets_2021_ICCV}, predicting semantic segmentations as additional inputs~\cite{Ye_2021_ICCV}, and crowdsourcing human demonstrations for imitation learning~\cite{habitatweb}).

\parbf{Performance}
For our experiments, we adopt the ``DD-PPO'' baseline, even though it scores very poorly on the leaderboard, and train our ImageNet and CLIP agents for 250M steps using DD-PPO. In~\cref{tab:habitat_objectnav}, we show that CLIP pretraining improves upon ImageNet pretraining by 1.45x SPL. Furthermore, our CLIP agent scores fourth place on the leaderboard, roughly on par with the leading entry and surpassing both winners of the 2021 challenge. This performance is particularly impressive because, unlike other entries, our agent does not (1) use depth information, (2) require additional data such as human annotations, or (3) require task-specific components. Furthermore, our agent has the second lowest distance to goal on the leaderboard, indicating that it navigates closer to the goal on average than most other models, even though it succeeds on fewer total episodes than the best models.

\subsection{Point Goal Navigation in Habitat}
\label{ssec:habitat_pointnav}

\begin{table}[t]
\centering

\begin{tabular}{lccc}
\toprule
Models & SPL & SR & Goal Dist \\
\midrule
\rowcolor{ColorTableUs}
ResNet-50 (CLIP) & \textbf{0.87} & \textbf{0.97} & \textbf{0.40} \\
ResNet-50 (ImageNet) & 0.82 & 0.94 & 0.73 \\
\bottomrule
\end{tabular}

\caption{
\textbf{Habitat \pnav.} As the 2019 leaderboard is now closed, we report full validation set metrics for our baselines.
}
\vspace{-0.05in}
\label{tab:habitat_pointnav}
\end{table}

\parbf{Task}
In \pnav, an agent must navigate from a random initial position to (relative) polar goal coordinates. We elect to provide agents with ideal GPS+Compass sensors (as in the Habitat 2019 challenge), as baseline agents are ineffective and hard to compare otherwise~\cite{Wijmans2020DDPPOLN}. The action space in \pnav is smaller than in \onav and includes just \texttt{MoveAhead}, \texttt{RotateRight}, and \texttt{RotateLeft}. The agent should call \texttt{Done} when it reaches its goal coordinates. We train only on the Gibson Database of 3D Spaces~\cite{xiazamirhe2018gibsonenv} (as permitted by the challenge).

\parbf{Prior work}
Prior methods made quick progress\footnote{\footnotesize\href{https://eval.ai/web/challenges/challenge-page/254/leaderboard/836}{\texttt{eval.ai/web/challenges/challenge-page\\/254/leaderboard/836}}} on \pnav with ideal sensors---\cite{Wijmans2020DDPPOLN} achieves 0.92 SPL on the Test Std set (when only RGB visual input is given), which is relatively high compared to the success metrics we see in other tasks (\cref{tab:robothor_objectnav,tab:ithor_rearrangement,tab:habitat_objectnav}). And, when provided with depth and additionally trained on the Matterport3D~\cite{Matterport3D} dataset, this same method achieves 0.94 SPL and 0.996 Success Rate on this set~\cite{Wijmans2020DDPPOLN}, so the task is regarded as ``solved''. This method consists of task-agnostic components and is trained for 2.5B steps, demonstrating continual improvements through this training period (although it achieves 90\% of total performance after 100M steps).

\parbf{Performance}
Because the ``DD-PPO'' agent has a task-agnostic design and this training method is now commonly used, we adopt it as the baseline for our experiment. We train the ImageNet and CLIP agents in our experiment for 250M steps using DD-PPO and report results in~\cref{tab:habitat_pointnav}. Again, our CLIP agent outperforms our ImageNet agent ($+0.05$ SPL and $+0.03$ Success Rate, with a 1.8x reduction in distance to goal). We observe that our model's performance is still increasing at the end of our training period. By the findings in Wijmans \emph{et al.}~\cite{Wijmans2020DDPPOLN}, we hypothesize that it could surpass the \cite{Wijmans2020DDPPOLN} 0.89 SPL leaderboard entry given the same duration of training. Our CLIP agent does also beat the 2019 challenge winner, which introduced ANM~\cite{chaplot2020learning}, by +0.14 SPL.

\section{Why is CLIP effective?}
\label{sec:why}

\begin{figure*}[tp]
    \centering
    \centering{
    \phantomsubcaption\label{fig:reachability}
      \phantomsubcaption\label{fig:object-presence-on-grid}
      \phantomsubcaption\label{fig:object-presence}
      \phantomsubcaption\label{fig:walkability}
      }
    \includegraphics[width=\textwidth]{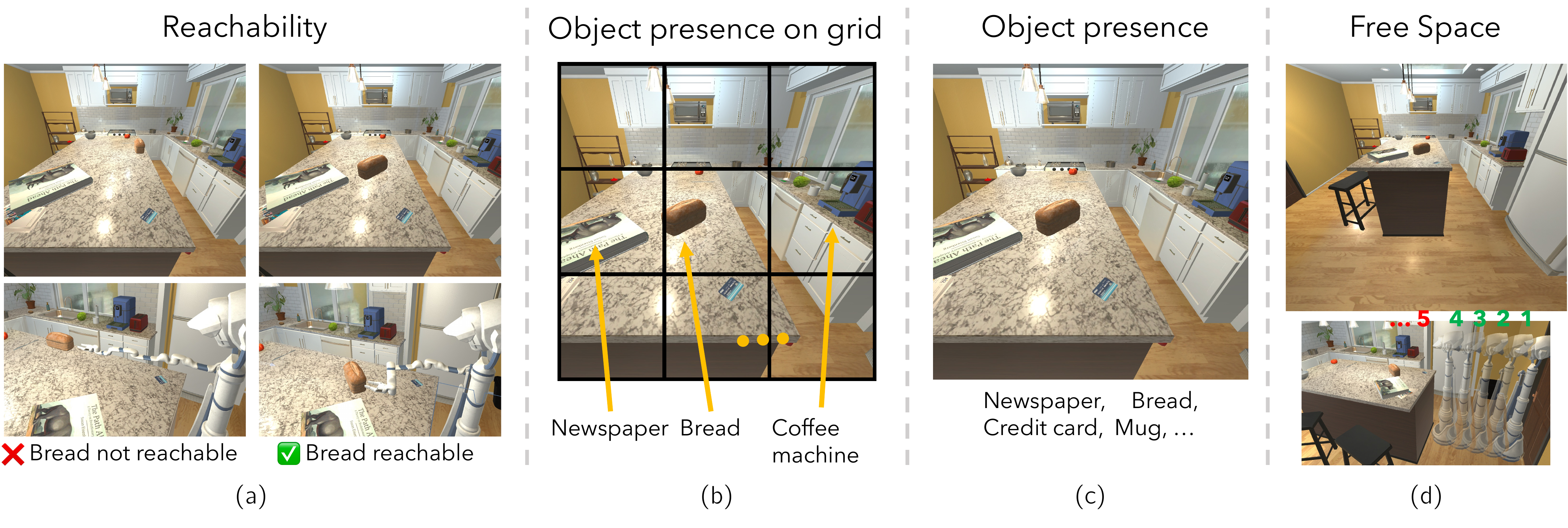}
    \caption{\textbf{Examples of visual encoder evaluations.} (a) Top: Two egocentric images, one in which the bread is reachable and one in which the bread is not reachable; bottom: two third-person images demonstrating the meaning of reachability (for visualization purposes only). (b) For each location on a $3{\times}3$ grid, the model must predict which objects are visible. (c) Predicting which objects are visible in an image in a multi-label fashion. (d) Top: image given to the model from which it must predict how many steps forward it can take before colliding with an object; bottom: third person viewpoint showing that the agent can, in this case, take 4 steps forward before colliding with the table.}
    \label{fig:linear-probe-examples}
    \vspace{0.1cm}
\end{figure*}

In the previous section, we showed that CLIP representations are remarkably effective across multiple embodied tasks, especially compared to their ImageNet counterparts. We wish to bring some insight as to \emph{why} this is the case.

To this end, we design a collection of experiments measuring how well visual representations from ImageNet and CLIP pretrained models encode the following semantic and geometric primitives: object presence, object presence at a location, object reachability, and free space. We posit that visual representations that encode these primitives effectively will lead to more capable Embodied AI agents, particularly given the nature of the 4 tasks we are investigating. 

For each of these primitives, we train simple (linear) classifiers to predict the outcome of interest from image features (with ResNet-50 encoders pretrained with either ImageNet or CLIP). See~\cref{fig:linear-probe-examples} for examples of these tasks. Our results, summarized in~\cref{tab:primitives}, show that classifiers trained on CLIP representations outperform on each of these tasks by values ranging from 3 to 6.5 points in comparison to those trained on ImageNet representations. Object localization shows a very large improvement (+6.5 absolute and +16\% relative improvement). Locating objects within an observation is useful for the two \onav tasks as well as the \roomr task. These results support the improvements we see in~\cref{sec:when}: they demonstrate that CLIP features natively encode information that is relevant to navigational and similar embodied skills.

To enable the above evaluation, we have generated a small dataset of frames containing objects from scenes in iTHOR. We select all 60 train, 15 val, and 15 test scenes from kitchens, living rooms, and bedrooms in iTHOR (and exclude bathrooms, which are too small). We sample 100 frames in each train scene and 50 frames from each val/test scene (from random locations and orientations). We additionally generate ground truth data per task (with details below) for each frame with the metadata available in iTHOR. We use this data for all tasks except reachability, for which we instead use the dataset and annotations from~\cite{Gadre2022ContinuousSR}.

For each frame in our datasets, we extract convolutional features using a ResNet-50 with CLIP or ImageNet pretraining and pool them into embeddings. At each training step, we classify them with a single linear layer and supervise using the ground truth data. We train with a batch size of 128, an Adam optimizer, and a learning rate of 0.001. We evaluate the best validation checkpoint for each model on our test set.

\begin{table}[t]
\centering
\footnotesize

\begin{tabular}{cllc}
\toprule
Task & Pretraining & Pooling & Score \\
\midrule
\multirow{3}{*}{Object Presence} & ImageNet & Average & 0.502 \\
& \textbf{CLIP} & \textbf{Average} & \textbf{0.530} \\
& CLIP & Attention & 0.529 \\
\midrule
\multirow{2}{*}{Object Localization} & ImageNet & Average & 0.387 \\
& \textbf{CLIP} & \textbf{Average} & \textbf{0.452} \\
\midrule
\multirow{3}{*}{Free Space} & ImageNet & Average & 0.287 \\
& \textbf{CLIP} & \textbf{Average} & \textbf{0.315} \\
& CLIP & Attention & 0.257 \\
\midrule
\multirow{3}{*}{Reachability} & ImageNet & Average & 0.638 \\
& \textbf{CLIP} & \textbf{Average} & \textbf{0.677} \\
& CLIP & Attention & 0.668 \\
\bottomrule
\end{tabular}

\caption{
\textbf{Do visual representations encode primitives?} We report test set scores for the experiments in~\cref{sec:why}. We compute F1 scores for Object Presence and Object Localization, and accuracy for Free Space and Reachability.
}
\vspace{-0.05in}
\label{tab:primitives}
\end{table}

\parbf{Pooling}
We use two methods to pool the $2048 \times 7 \times 7$ feature tensor created by the ResNet-50 encoder into a vector embedding. Note: pooling for ``object localization'' differs slightly, as described in the corresponding section below.

\noindent \emph{Average}: We average pool our conv. features from $7 \times 7$ to $1 \times 1$ and flatten, resulting in a 2048-dim embedding.

\noindent \emph{Attention}: We use the attention pooling mechanism from CLIP to produce a 1024-dim embedding. This is only applicable for features from a CLIP model and this module is not trained further in our experiments.

\parbf{Object presence}
In this task, we train our models to predict whether objects are present in the given image. See~\cref{fig:object-presence}. This task is not trivially solved by object detection, because objects in our images may be very small (\eg only a few pixels wide)---like in the embodied task of object navigation, our classifier needs to indicate whether an object is likely to be present ahead, even if it is not explicitly visible. We have selected a set of 52 object categories, so the linear layer has an output dimension of 52. We apply a Sigmoid activation on the outputs of this classifier to produce independent class probabilities $o=(o_0, \ldots, o_{51})$ (where $o_i > 0.5$ indicates that an object of category $i$ is present somewhere in the image). We supervise this with a binary cross-entropy loss.

\parbf{Object localization}
This task is quite similar to ``object presence'' (above) and uses the same set of 52 object categories. We divide the image into a $3 \times 3$ rectangular grid and the model must now predict whether objects are present in each of the grid sections (see~\cref{fig:object-presence-on-grid}). Spatial information about objects is highly relevant for navigational tasks (\eg ObjectNav), as this should help agents orient their vision and movement. We average pool our convolutional features from $7 \times 7$ to $3 \times 3$ (\ie corresponding to each grid section), instead of $1 \times 1$. We also disregard attention pooling of CLIP features here. During training, we then convolve these features with a $1 \times 1$ kernel from 2048 to 52 channels (\ie number of objects)---this convolution is equivalent to a single linear layer. We apply a Sigmoid activation and then have indicators for the presence of each object in each grid section. Our prediction is supervised with a binary cross-entropy loss.

\parbf{Free space}
In this task, we train our models to predict how many steps an agent can move forward (in 0.25m increments) given an egocentric view (see~\cref{fig:walkability}). Determining the amount of free space ahead is important for agents to plan where to move and how to avoid collisions. We design this as a classification problem for classes $c = \{0, \ldots, 9, \geq 10\}$, so the linear layer has an output dimension of 11. We apply a Softmax activation on the outputs of this classifier to produce probabilities and supervise these with a categorical cross-entropy loss.

\parbf{Reachability}
In this task, we train our models to predict whether an agent would be able to reach out and pick up some type of object from its current pose (see~\cref{fig:reachability}). This skill is especially relevant for embodied tasks like ArmPointNav in ManipulaTHOR~\cite{Ehsani2021ManipulaTHORA}, which involves picking up and moving objects with a robotic arm. Our dataset has 57 object categories and we have balanced the number of positive and negative examples per category (\ie whether an object is reachable when present). Our linear layer accordingly has an output dimension of 57 and we apply a Sigmoid activation on its outputs to produce independent class probabilities. During training, we iterate over each object in the set of all training images and supervise with a binary cross-entropy loss (between whether that object is indeed reachable and the model's predicted probability for that object class).

\section{Does ImageNet performance correlate with Embodied AI success?}
\label{sec:analysis}

\begin{figure}[tp]
    \centering
    \includegraphics[width=0.85\linewidth]{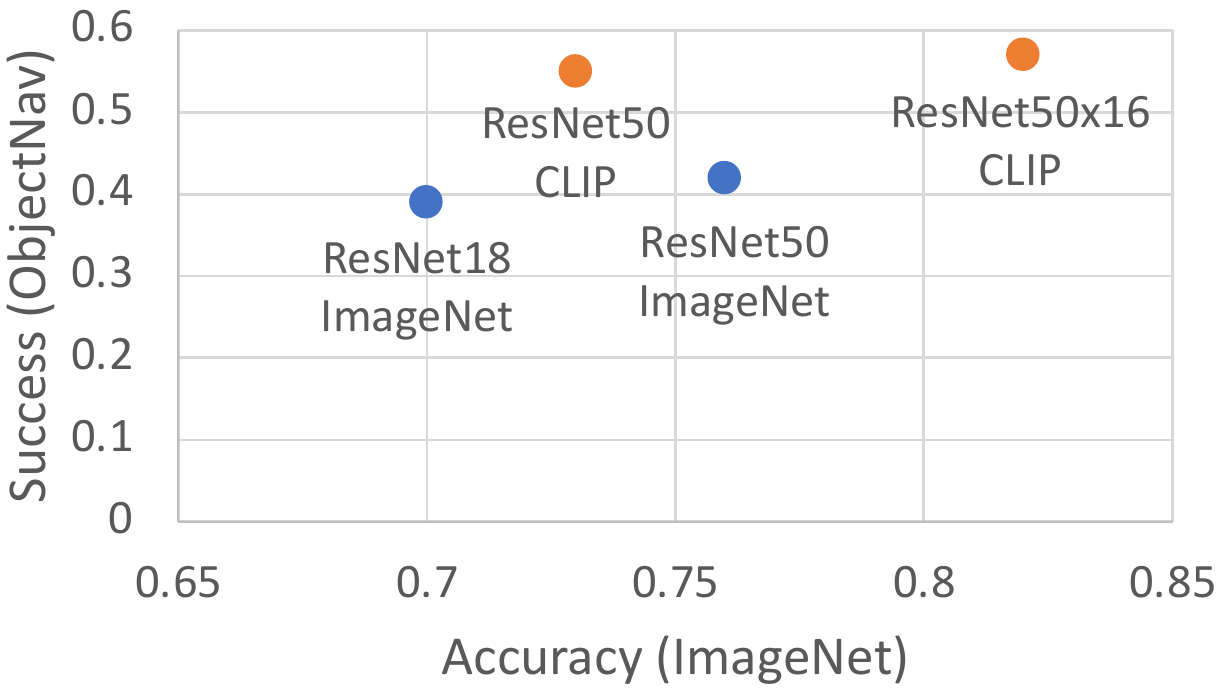}
    \caption{Comparison of \onav Success Rates by ImageNet accuracies for ResNet models with ImageNet or CLIP pretraining.}
    \label{fig:imagenet_vs_success}
\end{figure}

Our analysis in~\cref{sec:when} shows that using CLIP encoders provides large gains across a suite of benchmarks. Now, we ask: How do we choose an appropriate visual encoder when the next generation of networks get built? Is high accuracy on ImageNet sufficient to predict a high Success Rate on Embodied AI tasks?

To answer this question, we use 4 pretrained visual encoders and train \onav models using each of them. We use the same architecture for each, as described in~\cref{sec:clip}. The four encoders we use are: ResNet-18 and ResNet-50 pretrained on ImageNet, and ResNet-50 and ResNet-50x16 pretrained via CLIP. \cref{fig:imagenet_vs_success} shows a plot of \onav Success Rate (Test) vs ImageNet Top-1 Accuracy. Within the ImageNet models, larger models lead to an improved Top-1 Accuracy and also an improved Success Rate, but the gain in Success Rate is small. This is also true within the CLIP models. However, the Success Rate gains between the ImageNet and CLIP models is larger. When we consider the same architecture (ResNet-50), higher Top-1 Accuracy does not lead to a higher Success Rate.

These results and our probing experiments (\cref{sec:why}) suggest that Embodied AI is still limited by visual encoders and there seems to be room to improve merely by improving the visual encoding. However, one must probe visual representations for semantic and geometric information; moving forward, merely looking at ImageNet's Top-1 Accuracy alone is not a good indicator for success in Embodied AI.

\section{Discussion}
\label{sec:conclusion}

\parbf{Limitations}
We employed frozen backbones for our experiments. Developing robust training techniques to update the backbones while completing the task is an interesting direction to explore. Another limitation of the work is that our encoders are trained with visual and textual information. Incorporating interactive actions during representation learning can potentially lead to richer representations.

\begin{figure}[tp]
    \vspace{-10px}
    \centering
    \includegraphics[width=\linewidth]{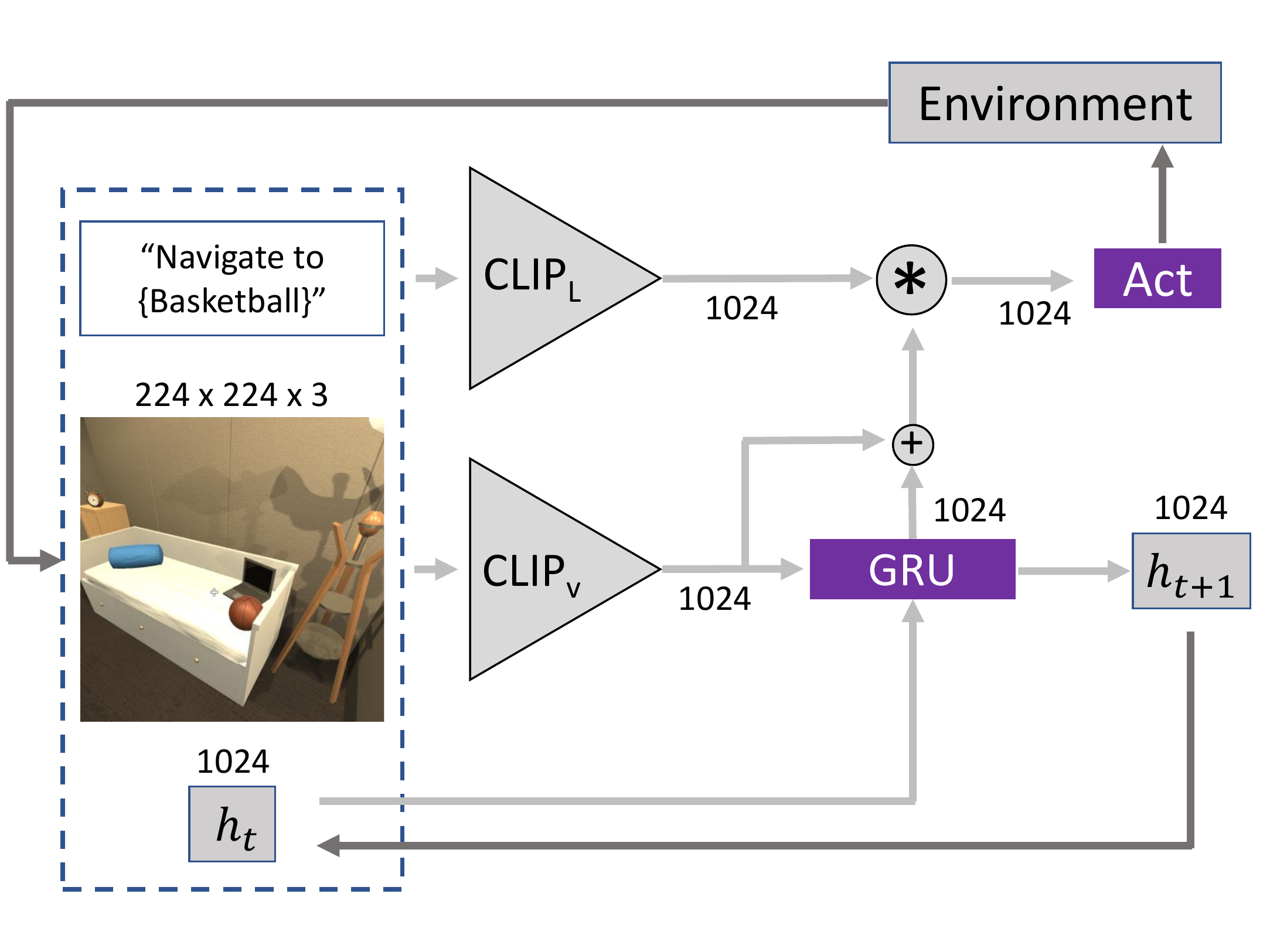}
    \vspace{-20px}
    \caption{\textbf{Zero-shot \onav\ Model Architecture.} This model has very few \colorbox{violet}{\textcolor{white}{learnable parameters}}, including the GRU, actor ($1024 \rightarrow 6$ linear layer), and critic ($1024 \rightarrow 1$ linear layer).
    }
    \label{fig:zeroshot}
\end{figure}

\begin{table}[t]
\centering
\resizebox{\columnwidth}{!}{%

\begin{tabular}{lcccccc}
\toprule
& Seen Objects & \multicolumn{5}{c}{Unseen Objects} \\
\cmidrule(lr){2-2} \cmidrule(lr){3-7}
Method & All & All & Apple & Basketball & House Plant & Television \\
\midrule
Random & 0.016 & 0.02 & 0.013 & 0.007 & 0.047 & 0.013 \\
Ours & \textbf{0.170} & \textbf{0.081} & \textbf{0.147} & \textbf{0.067} & \textbf{0.053} & \textbf{0.060} \\
\bottomrule
\end{tabular}

}
\caption{
\textbf{Zero-shot Success Rates for \onav.} We report metrics for 8 seen and 4 unseen objects in validation scenes.
}
\vspace{-0.05in}
\label{tab:zeroshot}
\end{table}

\parbf{Zero-shot \onav}
We are highly motivated by the effectiveness of CLIP's vision and language encoders in zero-shot image classification. As we've seen that CLIP produces meaningful visual representations for observations in an embodied environment (\cref{sec:why}), we are hopeful that \onav agents can learn using object goals encoded through the CLIP text encoder. We train an agent in RoboTHOR for 60M steps using DD-PPO with the simple baseline model in~\cref{fig:zeroshot}. This model has very few learnable parameters and operates entirely on CLIP representations as inputs. We train on 8 of the 12 objects and show success metrics from evaluation in~\cref{tab:zeroshot} for seen and unseen objects. On seen objects, this agent achieves a 0.17 Success Rate, while on unseen it achieves 0.08---roughly half. This is an interesting result, particularly when one considers the simplicity of the model used and the small number of episodes trained (just 60M) in comparison to other models in this paper and and on leaderboards. At 60M episodes, the agent achieves a 0.45 Success Rate for training scenes, so we expect this model to keep improving.

This is a good first step, and future work should build upon this result to explore architectures and training strategies that can improve zero-shot performance.

\parbf{Further considerations}
Our study opens up several very interesting questions for future work. In particular, because of CLIP's semantic training objective and distributions, it was a surprising finding that its image features do encode the reachability and free space primitives more effectively. Is contrastive learning or diversity of data the reason behind CLIP's overall effectiveness? We wonder if CLIP's distribution of training text might not encourage learning strong geometric priors, which are useful for navigation and determining free space. This might inspire the addition of explicit geometric objectives for future CLIP-style models. Finally, we are also interested by how the sparsity of CLIP features might encode higher-level navigational primitives.

\parbf{Conclusion}
We demonstrated the effectiveness of CLIP representations for navigation-heavy Embodied AI tasks, even beating specialized architectures at some challenges. To diagnose CLIP's effectiveness, we studied how well these representations encode relevant primitives for navigational tasks and found that CLIP models outperformed ImageNet counterparts. Finally, we also leveraged CLIP to enable an initial baseline for zero-shot \onav. We anticipate exciting future investigations into improved visual representations for embodied agents, as well as further progress on the zero-shot \onav task.

{\small
\bibliographystyle{ieee_fullname}
\bibliography{11_references}
}

\ifarxiv \clearpage \appendix
\label{sec:appendix}

\section{Baseline Architecture Details}

We provide further details here for implementations of baseline architectures for tasks in~\cref{sec:when}. We have already given such a description for RoboTHOR \onav in~\cref{sec:clip}. Each of these architectures is a replica of the baseline provided by task authors, with the visual encoder substituted with a frozen CLIP ResNet-50 encoder. And, for our ImageNet baseline agents, we simply use a frozen ResNet-50 that is pretrained on ImageNet instead.

\subsection{Room Rearrangement in iTHOR}

The model for \roomr in iTHOR receives two $3{\times}224{\times}224$ RGB images at every step $i_1$ and $i_2$. These RGB image are encoded into $2048{\times}7{\times}7$ tensors $I_1$ and $I_2$ by a CLIP ResNet-50 model whose weights are frozen and final attention pooling and classification layers have been removed. These feature maps are stacked to form a $6144{\times}7{\times}7$ tensor $s = [I_1, I_2, I_1 * I_2]$. An attention mask formed by applying a $1{\times}1$ convolution to $s$ with an output dimension of 512. $s$ is then convolved to 512 channels (resulting in a shape of $512{\times}7{\times}7$) with another $1{\times}1$ kernel. The attention mask is then used for attention pooling, resulting in a 512-dim embedding $V$. $V$ is passed into a 1-layer GRU with 512 hidden units, along with any prior hidden state. An actor head (one linear layer) maps the GRU output to a 6-dimensional vector of logits and a critic head (another linear layer) maps it to a scalar.

\subsection{Habitat \onav and \pnav}
We use the same architecture for our \pnav and \onav baselines in Habitat, with the only difference being the input goal and how it is encoded (to a 32-dim encoding $G$). Like in RoboTHOR, the input goal for Habitat \onav is an integer $g \in \{0,...,20\}$ indicating an object category. In this case, $g$ is used to index an embedding matrix and form $G$. In Habitat \pnav, the input goal is a 2-dim polar coordinate (to the target position, expressed relatively to the agent's current position). Here, $g$ is passed through a linear layer to form $G$. Weights for both the embedding matrix and linear layer are learned during training.

The model receives $G$, a $3{\times}224{\times}224$ RGB image $i$, the action from the previous step $a$ (as an integer index in the action space). The RGB image is encoded into a $2048{\times}7{\times}7$ tensor $I$ by a CLIP ResNet-50 model whose weights are frozen and final attention pooling and classification layers have been removed. $I$ is average pooled spatially (from $7{\times}7$ to $1{\times}1$) and flattened to form $V$: a 2048-dim visual embedding. The previous action $a$ is used to index an embedding matrix to form a 32-dim encoding $A$. $G$, $V$, and $A$ are passed into a 2-layer GRU with 512 hidden units, along with any prior hidden state. An actor head (one linear layer) maps the GRU output to a 6-dimensional vector of logits and a critic head (another linear layer) maps it to a scalar.
 \fi

\end{document}